\pdfoutput=1
\documentclass{article}
\usepackage[utf8]{inputenc}
\PassOptionsToPackage{numbers,sort&compress}{natbib}
\usepackage{main}
\usepackage{microtype}
\usepackage{graphicx}
\usepackage{times}
\usepackage{amsmath}
\usepackage{enumitem}
\usepackage{booktabs}
\usepackage{colortbl}
\usepackage{tabularx}
\usepackage{multirow}
\usepackage{wrapfig}
\usepackage{xcolor}
\definecolor{eoblue}{RGB}{0,84,160}

\definecolor{mydarkblue}{rgb}{0,0.08,0.45}
\usepackage[colorlinks=true,linkcolor=mydarkblue,citecolor=mydarkblue,filecolor=mydarkblue,urlcolor=mydarkblue]{hyperref}
\usepackage{fancyhdr}
\usepackage{xspace}
\hypersetup{
  pdftitle={The Last Remote Sensing Model Built by Humans? Casting Any Earth Observation Task as Recursive Self-Improvement},
  pdfauthor={Sijie Zhao, Zeqi Zhou, Heli Qi, Xuanchao Fu, Minghao Liu, Zhuo Zheng, Junshi Xia, Junjue Wang, Weihao Xuan, Naoto Yokoya}}
\usepackage[capitalise]{cleveref}
\crefname{section}{Sec.}{Secs.}
\Crefname{section}{Section}{Sections}
\crefname{table}{Tab.}{Tabs.}
\Crefname{table}{Table}{Tables}
\crefname{appendix}{App.}{Apps.}
\Crefname{appendix}{Appendix}{Appendices}

\newcommand{\method}{RSI4EO\xspace}

\newcommand{\tcite}[1]{~{\scriptsize\cite{#1}}}
\newcommand{\gain}[1]{\,{\scriptsize\textcolor{eoblue}{+#1}}}
\newcommand{\drop}[1]{\,{\scriptsize\textcolor{eoblue}{$-$#1}}}
\newcommand{\yr}[1]{\,{\fontsize{6}{7}\selectfont(#1)}}
\newcommand{\classic}[1]{\textcolor{gray}{#1}}
\newcolumntype{R}[1]{>{\raggedleft\arraybackslash}p{#1}}
\newcolumntype{L}[1]{>{\raggedright\arraybackslash}p{#1}}

\makeatletter
\renewcommand{\@toptitlebar}{
  \vskip -0.10in
  {\color{black}\hrule height 1\p@}
  \vskip 0.08in
  \vskip -\parskip%
}
\makeatother

\newenvironment{compactitem}%
 {\begin{itemize}[leftmargin=10pt,itemsep=0pt,parsep=0pt,topsep=2pt]}%
 {\end{itemize}}

\begin{document}

\title{The Last Remote Sensing Model Built by Humans? \\
Casting \textcolor{eoblue}{\textit{Any} Earth Observation Task} \\
as Recursive Self-Improvement}

\author{
\textbf{Sijie Zhao}$^{1*}$ \enspace
\textbf{Zeqi Zhou}$^{1*}$ \enspace
\textbf{Heli Qi}$^{2}$ \enspace
\textbf{Xuanchao Fu}$^{2}$ \enspace
\textbf{Minghao Liu}$^{1}$ \enspace
\textbf{Zhuo Zheng}$^{3}$ \enspace
\textbf{Junshi Xia}$^{2}$ \\
\textbf{Junjue Wang}$^{1,2}$ \enspace
\textbf{Weihao Xuan}$^{1,2\dagger}$ \enspace
\textbf{Naoto Yokoya}$^{1,2\dagger}$ \\
\textsuperscript{1}The University of Tokyo \quad
\textsuperscript{2}RIKEN \quad
\textsuperscript{3}Stanford University \\
\href{https://github.com/Yokoya-Lab-UTokyo/RSI4EO}{\includegraphics[height=0.4cm]{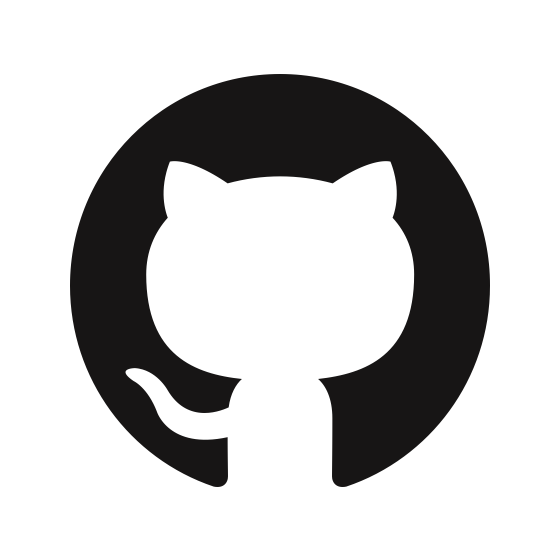} \textbf{Yokoya-Lab-UTokyo/RSI4EO}} \qquad
\href{https://huggingface.co/UTokyo-Yokoya-Lab/RSI4EO}{\raisebox{-0.05cm}{\includegraphics[height=0.4cm]{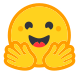}} \textbf{UTokyo-Yokoya-Lab/RSI4EO}}
}

\maketitle
{\let\thefootnote\relax\footnotetext{$^{*}$Equal contribution. \quad $^{\dagger}$Corresponding authors.}}
\pagestyle{plain}
\thispagestyle{fancy}
\fancyhead{}
\lhead{%
  \includegraphics[height=0.56cm]{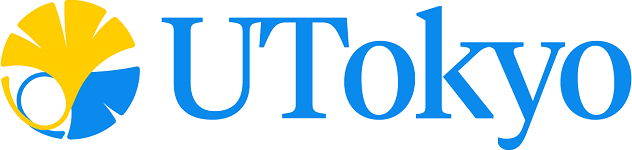}%
  \hspace{0.35cm}%
  \includegraphics[height=0.775cm]{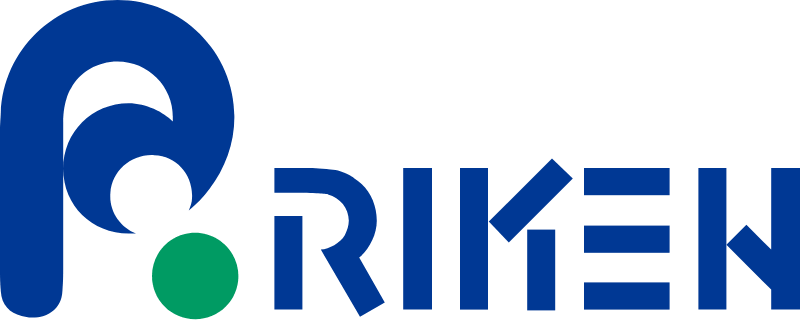}%
}
\rhead{}
\renewcommand{\headrulewidth}{0pt}
\setlength{\headheight}{23.3pt}
\addtolength{\topmargin}{-11.3pt}
\setlength{\headsep}{1.5mm}

\vspace{-1.0em}

\begin{center}
\includegraphics[width=\linewidth]{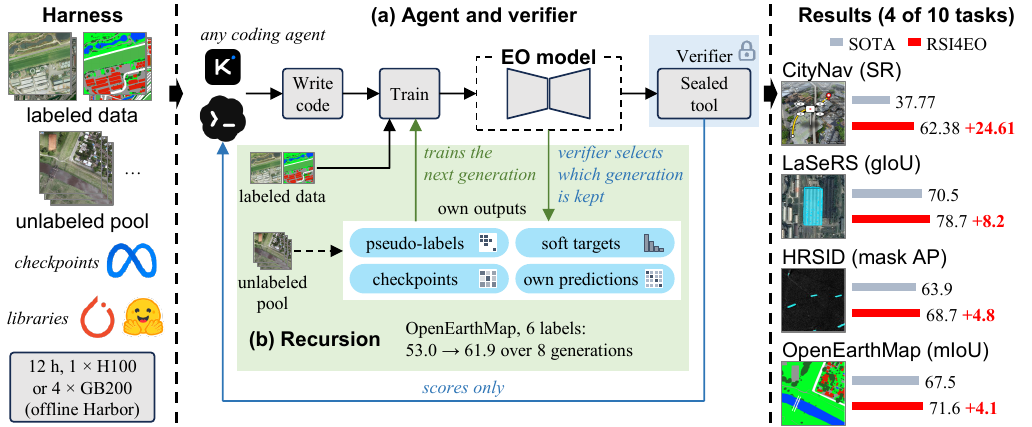}
\captionof{figure}{\method casts any EO task as a harness and a verifier. (a)~A general-purpose coding agent writes and trains the EO model and receives only the scores of the sealed verifier. (b)~Each generation of the model is built from the outputs of the previous one, and the verifier decides which generation is kept. The resulting models surpass the state of the art on all 10 tasks (\cref{tab:main}).}
\label{fig:teaser}
\end{center}

\begin{abstract}
Earth observation (EO) models are still built by hand. For each new task, experts design a training recipe and tune it against a benchmark. We show that this loop can be handed to an agent. \textbf{Any verifiable task can be cast as a harness and a verifier}, where the harness defines what the agent may use and the verifier defines what counts as success. Given only these two components, a general-purpose coding agent builds the EO model without human input and improves it recursively on its own outputs. The many sources of EO ground truth become verifiers, and settings long treated as separate research problems, from learning with a handful of labels to responding to a new disaster, become choices of the two components. On 10 EO tasks of five types, models that an agent builds within 12 hours surpass the state of the art on every one, against more than 1,000 published methods. The median lead is about half of all the progress that these methods have made over the classic baseline. With 0.2 to 4.3\% of the training labels, agent-built models reach 72 to 108\% of the state-of-the-art score obtained with all labels, and 10\% of all labels, those of the verifier included, surpass the state of the art on land-cover mapping. On a single GPU, an agent surpasses it on eight of the 10 tasks, and stronger agents build stronger models. \textbf{This establishes a new paradigm for EO, in which people design harnesses and verifiers and agents build the models}, so EO models improve as agents do.
\end{abstract}

\section{Introduction}
\label{sec:intro}

Coding agents now carry out the engineering behind a model for hours, writing, training and debugging code against a validation metric \cite{jiang2025aide,chan2025mlebench,karpathy2026autoresearch}. On self-contained machine-learning tasks, they already reach medal-level scores in Kaggle competitions and outscore human experts given the same two hours \cite{chan2025mlebench,wijk2025rebench}. Whether a model built this way can surpass what a research community has built over years remains an open question. The answer decides whether people keep building models by hand or hand the task to an agent.

Earth observation (EO) is a demanding and consequential setting for this question. Its data span optical and SAR imagery, multispectral time series and weather radar, and its tasks range from land-cover mapping to rainfall nowcasting and drone navigation. With more than 500 public datasets \cite{xiong2022earthnets}, EO provides a solid basis for testing how well and how robustly an agent builds models across tasks. The same diversity keeps EO reliant on models that experts build for each task. Foundation models and multimodal large models raise the starting point \cite{xiao2025rsfmsurvey} but do not consistently outperform a supervised baseline \cite{marsocci2026pangaea,fogel2025opencanopy}. Where these models do help, experts still adapt them to each task \cite{zhang2025vicrop,zhang2026gazeearth}. As a result, each benchmark has been pushed by dozens to hundreds of specialized methods, a demanding reference for a model built by an agent. EO also needs models built automatically. A specialized model takes weeks to months of expert engineering, whereas a new disaster needs one within days, before a training set exists. In 2024 alone, the International Charter on Space and Major Disasters was activated 85 times in 52 countries \cite{charter2024}. For many agencies, countries and regions, building a specialized model remains out of reach \cite{rolf2021mosaiks}. Agents built for EO have so far orchestrated pretrained models and tools or refined fully supervised baselines \cite{feng2026earthagent,kang2026ringmoclaw}. No agent has yet been handed a whole EO task and its environment to build the model itself.

To this end, we propose \method (Recursive Self-Improvement for EO), which casts any verifiable EO task as a harness and a verifier. AI has advanced fastest wherever success can be checked automatically \cite{guo2025deepseek,novikov2025alphaevolve}, and the two components give an EO task exactly this property (\cref{fig:teaser}). The harness defines what the agent may use, such as data, pretrained checkpoints, libraries and compute. The verifier defines what counts as success. It scores every model the agent submits on data the agent cannot see and returns only the score. Within this environment, a general-purpose coding agent builds the model offline and without human input \cite{harbor2026framework,merrill2026terminalbench}. It proposes and implements changes to the model, the verifier decides which of the resulting candidates to keep, and each kept model and its outputs become the starting point of the next change. The model improves recursively on its own outputs \cite{zelikman2024stop}, and the agent designs this recursion itself. We design the task and leave the agent as it is. The harness and the verifier are the only parts written for EO, so any general-purpose coding agent can take the task. \method thereby unifies the diverse tasks and learning settings of EO. The many sources of EO ground truth become verifiers, and learning from few labels, training-free adaptation and domain adaptation to a new disaster become choices of the two components (\cref{sec:settings}).

Within 12 hours, an agent equipped this way builds models that surpass the state of the art on all 10 tasks of five types (\cref{tab:benchmarks,tab:main}), against more than 1,000 published methods. The margins are large (\cref{sec:results}). Across the 10 tasks, the median lead over the previous state of the art is about half of all the progress that published methods have made over the classic baseline, and on drone navigation by language, the success rate rises from 37.8 to 62.4\%. The lead holds even where the harness forbids training or offers no pretrained model for the task. A few training labels already recover most of this performance (\cref{sec:labels}). With 0.2 to 4.3\% of the training labels, agent-built models reach 72 to 108\% of the state-of-the-art score obtained with all labels. Strong models also come from a single GPU and from another developer's agent. On a single H100 GPU, GPT-6.1-Sol surpasses the state of the art on 8 of the 10 tasks, and Kimi-K3.1-Pro, from another developer, on 6.

Two controlled comparisons show that a verifier of a few labels suffices and that the model improves with the agent. The first varies the number of labels and their split between training and the verifier, on land-cover mapping and on an earthquake held out from training (\cref{sec:labels}). A verifier needs only a few labels, and the rest serve best in training, which we call the train-first rule. Split this way, 10\% of the land-cover labels surpass the previous state of the art, and a third of the earthquake's labels surpass a model trained on all of them. Without a single label of the earthquake, the agent already surpasses every published zero-shot, domain-adaptation and semi-supervised result. The second comparison fixes the harness and the verifier and varies only the agent (\cref{sec:agents}). From the oldest to the newest of three generations, the models of the GPT and Kimi families gain 7.9 and 8.5 mIoU on six-label land-cover mapping.\looseness=-1

The trajectories show why agent-built models surpass the published state of the art: the agent designs the whole system around each task, and the verifier keeps the candidates that improve the model (\cref{sec:analysis}). The design reaches from how the data are read to how the predictions are combined, and its choices differ with the task. For instance, the agent keeps a pretrained encoder for land-cover mapping but trains temporal models from scratch for nowcasting. The gain comes from this system and not only from the pretrained models. Against published methods that build on the pretrained models of the harness, the agent-built models lead on all eight tasks compared, by 2.0 to 32.6 points and by 0.45\,m of canopy-height error. From the many candidates that the agent proposes, the verifier keeps those that improve the model. Over eight generations of self-training on land-cover mapping with six labeled tiles, each kept model labels the unlabeled tiles for the next, and the agent finally writes scripts that run this loop without its intervention. On the same task, removing the scores of the verifier or the recursion leaves a model 21.6 and 5.3 mIoU weaker. The models selected this way do not overfit the verifier. It ranks the models of nine agents exactly as the hidden test does, and hidden-test scores stay within 2.5 points of verifier scores in every run on land-cover mapping and question answering.

These results establish a new paradigm for EO, in which people design harnesses and verifiers, and agents build the models. A task's harness and verifier stay fixed while agents improve \cite{kwa2025metr,chan2025mlebench}, so EO models improve as agents do, without new human engineering. A new disaster, region or satellite receives a model within hours rather than after weeks of engineering, and any agency or country that can define a task can obtain a model at the level of the published state of the art. Research effort moves upstream to which verifiers capture a task faithfully and what a harness must contain. Our contributions are threefold.
\begin{compactitem}
\item \textbf{A Paradigm for Building EO Models.} We cast any verifiable EO task as a harness and a verifier, with which a general-purpose coding agent builds the model and improves it recursively. The many sources of EO ground truth become verifiers, and learning settings studied separately become choices of the two components.
\item \textbf{An Open Task Suite for the Community.} We release 10 EO tasks of five types as harness and verifier pairs that run in offline Harbor containers. Any researcher can hand these tasks to their own agent, compare the resulting models with published methods and with ours, and add new tasks in the same form.
\item \textbf{Evidence and Insights.} Agent-built models surpass the state of the art on all 10 tasks (\cref{sec:results}), against more than 1,000 published methods. With 0.2 to 4.3\% of the training labels, they reach 72 to 108\% of the state-of-the-art score, and the split of labels follows a train-first rule (\cref{sec:labels}). The model improves with the agent (\cref{sec:agents}), and the trajectories and two ablations trace the gains to a system that the agent designs around each task and the verifier selects (\cref{sec:analysis}).
\end{compactitem}

\section{Related Work}
\label{sec:related}

\paragraph{Recursive Self-Improvement and Verifiers.}
Machines that design better machines were formalized in the G\"odel machine \cite{schmidhuber2007godel}, and language models now realize parts of it. STOP improves the scaffold that calls it \cite{zelikman2024stop}, the Darwin G\"odel Machine rewrites its own code \cite{zhang2026darwin}, and AlphaEvolve evolves programs against automatic evaluators \cite{novikov2025alphaevolve}. Self-training lets a model improve on its own outputs \cite{xie2020noisy,zelikman2022star}, and progress here follows the availability of verifiers, up to reinforcement learning with verifiable rewards \cite{guo2025deepseek,chen2026rsi}. In these works, people fix which artifact improves and how each round learns from the last. In \method, the artifact that improves is the EO model itself, and the agent designs the recursion, including how each generation learns from the last.\looseness=-1

\paragraph{Agents for Machine Learning Engineering.}
AutoML searches spaces that people define \cite{isensee2021nnunet}, and LLM agents extend this space to arbitrary code. AIDE, The AI Scientist and autoresearch revise training code against validation metrics \cite{jiang2025aide,lu2026aiscientist,karpathy2026autoresearch,aygun2026era,panfilov2026claudini}, and MLE-bench, RE-Bench and PostTrainBench evaluate these agents \cite{chan2025mlebench,wijk2025rebench,rank2026posttrainbench}. MLE-bench and RE-Bench grade them against competition leaderboards and human experts at work for hours, whereas we grade agent-built models against the published state of the art of research benchmarks. These works ask how well an agent solves a given task, and EurekAgent argues that the bottleneck is shifting from prescribing agent workflows to designing agent environments \cite{xin2026eurekagent}. \method makes the task itself the object of design: a harness and a sealed verifier specify any EO task, and learning settings studied as separate problems become choices of the two. Sealing the verifier follows reward-hacking studies, which find agents training on the test set \cite{rank2026posttrainbench}.

\paragraph{Earth Observation Models and Agents.}
EO foundation models and satellite-pretrained backbones raise the starting point of EO models \cite{jakubik2025terramind,simeoni2025dinov3}, and label-scarce and cross-domain settings have each produced dedicated methods \cite{yang2025unimatchv2,hoyer2023mic}. In \method, these models and methods become ingredients that the agent selects, combines and trains. LLM agents for EO orchestrate fixed tools and pretrained models or create their own tools \cite{feng2026earthagent,zhao2026openearth}, and two recent systems retrain models to improve fully supervised baselines \cite{kang2026ringmoclaw,lekavicius2026solar}. \method hands the agent a whole EO task and its environment, and the agent-built models reach beyond fully supervised baselines, to few or no target labels and to photogrammetry, nowcasting and closed-loop navigation.

\section{Method}
\label{sec:method}

\subsection{Tasks as a Harness and a Verifier}
\label{sec:task}

An EO task is a pair $\mathcal{T}=(\mathcal{H},\mathcal{V})$. The harness $\mathcal{H}$ fixes the resources the agent may use: labeled data $\mathcal{D}_L$, unlabeled data $\mathcal{D}_U$, pretrained checkpoints, software and a compute budget. The verifier $\mathcal{V}$ fixes what counts as success. The agent produces pipelines, where a pipeline $\pi$ is a self-contained directory of code and trained weights that maps input imagery to predictions. The verifier runs a pipeline on held-out data $(X_V,Y_V)$ that the agent never sees and returns its score
\begin{equation}
s(\pi)=m\big(\pi(X_V),\,Y_V\big),
\end{equation}
where $m$ is the task metric. Over its budget, the agent writes pipelines $\pi_1,\dots,\pi_K$, queries $s(\pi_k)$ and decides what to try next. When the agent stops or the budget ends, its final pipeline $\pi^\star$ is evaluated once on a hidden test set with the same metric, $m(\pi^\star(X_T),Y_T)$.

Every modeling decision, from the architecture to the use of unlabeled data, belongs to the agent's pipeline. A task is fully specified by its two components, so any agent can receive it, and changing either component defines a new task.

\subsection{Verifier}
\label{sec:verifier}

\begin{figure}[!t]
\centering
\begin{minipage}[t]{0.485\linewidth}
\centering
\includegraphics[width=\linewidth]{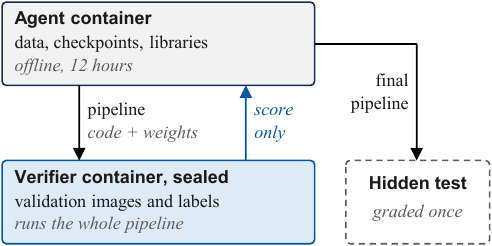}
\caption{The verifier is sealed. A pipeline goes in and only its score comes back. The hidden test grades the final pipeline once.}
\label{fig:sealed}
\end{minipage}\hfill
\begin{minipage}[t]{0.485\linewidth}
\centering
\includegraphics[width=\linewidth]{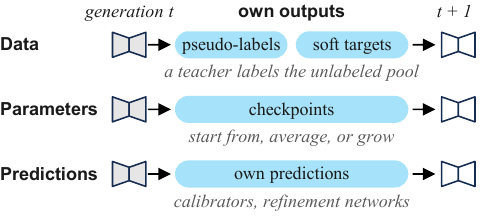}
\caption{Each generation learns from the last at three levels. The agent chooses the mechanisms, their order and the depth of the lineage.}
\label{fig:mechanisms}
\end{minipage}
\end{figure}

The verifier is sealed (\cref{fig:sealed}). It runs in a separate container that holds the validation data, executes a submitted pipeline exactly as the final grader will, and returns the score through a tool call. The agent never receives validation images or labels. Because the verifier runs the whole pipeline, it scores what will be graded, including preprocessing, ensembling and test-time augmentation. The agent may query the verifier as often as it likes, and the scores guide its next candidates.

EO is rich in sources for such verifiers: held-out sets of labeled benchmarks, independent sensors and in-situ measurements, physical models, and the repeated observation of the same place. Our verifiers already reach beyond human annotation. Six hold annotated benchmark data. The canopy-height and stereo verifiers score against airborne LiDAR, an independent sensor, and the nowcasting verifier scores each forecast against the radar frames observed afterwards, which needs no annotation. The navigation verifier flies each policy in closed loop in a simulator of held-out city blocks.

\subsection{Harness}
\label{sec:harness}

The harness gives the agent what an EO researcher would start from: the labeled training data, an unlabeled pool where the task provides one, the strongest general-purpose pretrained models that suit the task, and the Python deep-learning and geospatial stack. The pretrained models include the DINOv2 and the satellite-pretrained DINOv3 encoders \cite{oquab2024dinov2,simeoni2025dinov3} and the vision--language model Qwen3-VL \cite{bai2025qwen3vl}. The starter pipeline is deliberately plain, so that the final model reflects what the agent builds rather than what it is given. \Cref{app:tasks} lists the models, tools and starter of every task.

The submitted pipeline must contain networks that the agent trained, except on the training-free task, where no parameter may be updated. The given labels may not be edited, and any further labels must come from code and models in the container, not from the agent inspecting images, and a scored pipeline may not learn at test time (\cref{app:agentsetup}).

Each task runs in an offline Harbor container \cite{harbor2026framework,merrill2026terminalbench}, where the agent works without human input or network access beyond its model interface. Each agent has 12 hours on four GB200 GPUs or on a single H100 GPU (\cref{sec:setup}).

\subsection{Recursive Self-Improvement}
\label{sec:recursion}

Within a task, the agent improves the EO model recursively on its own outputs. We call each model that the agent trains from the outputs of earlier models a generation. A generation $f_t$ leaves outputs $O_t$, including its predictions and class probabilities on the unlabeled pool, its checkpoints, and its predictions on held-out folds of the training data. The next generation is trained from these outputs,
\begin{equation}
f_{t+1}=\mathrm{Train}\big(\mathcal{H},\,O_{\le t}\big),
\end{equation}
and the verifier decides which generation is kept as the source of the next one. Two loops run together. In the outer loop, the agent proposes pipelines and the verifier scores them. In the inner loop, each kept model becomes material for its successor, so the model improves along a lineage (\cref{fig:teaser}).

The agent designs this recursion itself. The task instruction permits methods such as pseudo-labeling on the unlabeled pool but prescribes none, and the agent chooses the mechanisms, their order and the depth of the lineage from the verifier's feedback, so the recursion is fitted to each task rather than fixed in advance. Across our runs, the mechanisms fall into three families (\cref{fig:mechanisms}). At the data level, a teacher labels the unlabeled pool with hard or soft labels and a student trains on them \cite{xie2020noisy,hinton2015distilling}. At the parameter level, new models start from earlier checkpoints, average several checkpoints, or grow from a smaller trained network. At the prediction level, the agent trains calibrators and refinement networks on its own predictions.

\subsection{Learning Settings as Choices of Harness and Verifier}
\label{sec:settings}

\begin{table}[!t]
\centering
\caption{The 10 tasks of five types. LRS-GRO, LaSeRS and CityNav count questions, queries and episodes (\cref{app:tasks}).}
\label{tab:benchmarks}
\fontsize{8}{9.8}\selectfont
\setlength{\tabcolsep}{3.4pt}
\begin{tabularx}{\textwidth}{@{}lXllrr@{}}
\toprule
Task & Input & Output & Metric & Train / Verifier / Test & Few-label task \\
\midrule
\multicolumn{6}{@{}l}{\textit{Image interpretation}} \\
\hspace{0.5em}OpenEarthMap \cite{xia2023openearthmap} & RGB, 0.25--0.5\,m & land-cover map & mIoU & 2,700 / 300 / 500 & 6 tiles (0.2\%) \\
\hspace{0.5em}BRIGHT \cite{chen2025bright} & optical + SAR pairs & damage map & mIoU & 2,918 / 450 / 878 & 14 pairs (0.5\%) \\
\hspace{0.5em}HRSID \cite{wei2020hrsid} & SAR, 0.5--3\,m & ship instances & mask AP & 3,090 / 552 / 1,962 & 36 images (1.2\%) \\
\hspace{0.5em}PASTIS \cite{garnot2021pastis} & Sentinel-2 time series & crop-type map & mIoU & 1,455 / 482 / 496 & 20 series (1.4\%) \\
\addlinespace[3pt]
\multicolumn{6}{@{}l}{\textit{Quantitative retrieval}} \\
\hspace{0.5em}Open-Canopy \cite{fogel2025opencanopy} & SPOT~6--7 RGB + NIR, 1.5\,m & canopy height & MAE (m) & 66,339 / 7,369 / 13,675 & 1,000 tiles (1.5\%) \\
\addlinespace[3pt]
\multicolumn{6}{@{}l}{\textit{Photogrammetry}} \\
\hspace{0.5em}WHU-Stereo \cite{li2023whustereo} & Gaofen-7 panchromatic stereo pairs & disparity & D1 (\%) & 1,220 / 122 / 415 & -- \\
\addlinespace[3pt]
\multicolumn{6}{@{}l}{\textit{Spatiotemporal forecasting}} \\
\hspace{0.5em}SEVIR \cite{veillette2020sevir} & weather-radar VIL, past 65\,min & next-hour echoes & CSI-M & 35,718 / 9,060 / 12,159 & -- \\
\addlinespace[3pt]
\multicolumn{6}{@{}l}{\textit{Vision--language}} \\
\hspace{0.5em}LRS-GRO \cite{liu2026zoomearth} & RGB, up to 27,328\,px, + question & answer & accuracy & 2,656 / 855 / 9,734 & 114 answers (4.3\%) \\
\hspace{0.5em}LaSeRS \cite{segearthr2} & RGB + free-form query & ordered masks & gIoU & 23,939 / 1,600 / 1,158 & -- \\
\hspace{0.5em}CityNav \cite{lee2025citynav} & orthophoto, DSM, landmarks + text & flight policy & SR (\%) & 22,002 / 2,826 / 5,311 & -- \\
\bottomrule
\end{tabularx}
\end{table}

\begin{wrapfigure}{r}{0.47\linewidth}
\centering
\vspace{-0.9em}
\includegraphics[width=\linewidth]{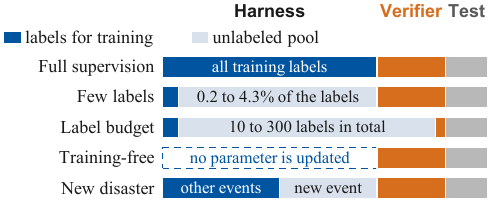}
\caption{Learning settings are choices of the harness and the verifier. Each row places the labels of a task differently.}
\label{fig:settings}
\vspace{-0.8em}
\end{wrapfigure}

\method unifies the diverse tasks and learning settings of EO as choices of the two components (\cref{fig:settings}). A task's inputs, outputs and source of ground truth fix what the harness holds and how the verifier scores, so land-cover mapping, stereo matching, nowcasting and drone navigation share one interface (\cref{tab:benchmarks}). Learning settings that EO research has treated as separate problems are set the same way. Full supervision gives the harness all training labels. Learning from few labels keeps a few labeled samples per natural stratum of the data, such as a continent or a disaster event, and moves the remaining training imagery to the unlabeled pool (\cref{app:tasks}). Training-free adaptation forbids training in the harness, so the agent can only design how a fixed foundation model is used. A label budget fixes how many labels a task holds and how they split between the harness and the verifier (\cref{sec:labels}). Domain adaptation places the labeled data on a source domain and the verifier on the target domain, and a new disaster is one such domain. When the target holds no labels at all, the verifier holds samples of the source domain instead. In every case, the agent, the verifier interface and the final evaluation stay the same.

\section{Experiments}
\label{sec:experiment}

The experiments establish three results: agent-built models surpass the state of the art on all 10 tasks (\cref{sec:results}), a handful of labels suffices (\cref{sec:labels}), and stronger agents build stronger models (\cref{sec:agents}). \Cref{sec:analysis} then examines why.

\subsection{Setup}
\label{sec:setup}

\paragraph{Tasks.} We instantiate \method on 10 tasks of five types that differ in sensor, output and source of ground truth (\cref{tab:benchmarks}). Every task yields a full-label task, and six also yield a few-label task with 0.2 to 4.3\% of the training labels while the verifier keeps its validation split. The hidden test set is the split on which published methods report their results, the official validation split on OpenEarthMap, whose 1,500 official test images join the unlabeled pool (\cref{app:tasks}).

\paragraph{Label Budgets.} The label-budget tasks cap all labels of a task, those of training and of the verifier together. They hold 0, 10, 30, 100 or 300 target labels on OpenEarthMap and on the 2023 Turkey earthquake, held out from BRIGHT \cite{chen2025bright} as a new disaster, and the remaining target imagery outside the test set joins the unlabeled pool. On BRIGHT, the agent also trains on the labels of the other 13 events. Three splits give nine tenths of the labels to training (train-first), half (even) or all of them to the verifier (verifier-only) (\cref{app:budgets}).

\paragraph{Agents and Protocol.} Three agents run every full-label task and the six few-label tasks: GPT-6-Astra in Codex on four GB200 GPUs, and GPT-6.1-Sol in Codex and Kimi-K3.1-Pro in Kimi Code, each on a single H100 GPU. All agents use their highest reasoning effort and work for at most 12 hours in an offline container, and each agent score in the tables and figures comes from a single run (\cref{app:agentsetup}).

\paragraph{Reference Points.} We compare each agent-built model with the starter of the harness, run as provided, and with published results under each task's standard protocol, among them the classic baseline and the previous state of the art of each task (\cref{tab:main,app:published}).

\subsection{Agent-Built Models Surpass the State of the Art on 10 Tasks}
\label{sec:results}
\begin{table}[!tp]
\centering
\caption{Agent-built models surpass the state of the art on all 10 tasks. Each panel lists six published results, with the classic baseline in grey and the previous state of the art (SOTA) underlined, both with their years, above the hidden-test scores of the starter and of the models built by three agents (blue rows) on the full-label and the few-label task. Beside each task name are its metric and the number of methods that report results on it. Bold scores are beyond the SOTA, with the margin.}
\label{tab:main}
\fontsize{8.5}{9.6}\selectfont
\setlength{\tabcolsep}{2pt}
\setlength{\aboverulesep}{0.2ex}
\setlength{\belowrulesep}{0.3ex}
\begin{minipage}[t]{0.485\textwidth}
\begin{tabularx}{\linewidth}[t]{@{}XR{25pt}L{24pt}R{23pt}L{16pt}@{}}
\multicolumn{5}{@{}l@{}}{\textbf{(a) OpenEarthMap}\enspace{\fontsize{7.5}{8.5}\selectfont\textcolor{gray}{mIoU, 256 methods}}} \\
\toprule
{\fontsize{7.5}{8.5}\selectfont\textcolor{gray}{\textit{Share of the training labels}}} & 100\% & & 0.2\% & \\
\midrule
\classic{UNet}\tcite{ronneberger2015unet}\classic{\yr{2015}} & \classic{57.8} & & & \\
RS3Mamba\tcite{ma2024rs3mamba} & 60.5 & & & \\
DC-Swin\tcite{wang2022dcswin} & 63.6 & & & \\
UNetFormer\tcite{wang2022unetformer} & 64.3 & & & \\
MetaSegNet\tcite{wang2024metasegnet} & 67.0 & & & \\
PyramidMamba\tcite{wang2025pyramidmamba}\yr{2024} & \underline{67.5} & & & \\
\midrule
Starter & -- &  & 34.0 &  \\
\rowcolor{eoblue!7}\method{} (GPT-6-Astra) & \textbf{71.6} & \gain{4.1} & 60.0 &  \\
\rowcolor{eoblue!7}\method{} (GPT-6.1-Sol) & \textbf{68.8} & \gain{1.3} & 59.6 &  \\
\rowcolor{eoblue!7}\method{} (Kimi-K3.1-Pro) & \textbf{69.7} & \gain{2.2} & 57.2 &  \\
\bottomrule
\end{tabularx}
\end{minipage}\hfill
\begin{minipage}[t]{0.485\textwidth}
\begin{tabularx}{\linewidth}[t]{@{}XR{25pt}L{24pt}R{23pt}L{16pt}@{}}
\multicolumn{5}{@{}l@{}}{\textbf{(b) BRIGHT}\enspace{\fontsize{7.5}{8.5}\selectfont\textcolor{gray}{mIoU, 127 methods}}} \\
\toprule
{\fontsize{7.5}{8.5}\selectfont\textcolor{gray}{\textit{Share of the training labels}}} & 100\% & & 0.5\% & \\
\midrule
\classic{UNet}\tcite{ronneberger2015unet}\classic{\yr{2015}} & \classic{64.94} & & & \\
SiamCRNN\tcite{chen2020siamcrnn} & 65.73 & & & \\
ChangeOS\tcite{zheng2021changeos} & 65.98 & & & \\
DamageFormer\tcite{chen2022damageformer} & 67.09 & & & \\
ChangeMamba\tcite{chen2024changemamba} & 67.63 & & & \\
Ensemble\tcite{chen2025bright}\yr{2025} & \underline{68.86} & & & \\
\midrule
Starter & 61.79 &  & 37.29 &  \\
\rowcolor{eoblue!7}\method{} (GPT-6-Astra) & \textbf{71.56} & \gain{2.70} & 49.42 &  \\
\rowcolor{eoblue!7}\method{} (GPT-6.1-Sol) & 66.60 &  & 39.21 &  \\
\rowcolor{eoblue!7}\method{} (Kimi-K3.1-Pro) & \textbf{69.02} & \gain{0.16} & 40.36 &  \\
\bottomrule
\end{tabularx}
\end{minipage}

\vspace{3pt}

\begin{minipage}[t]{0.485\textwidth}
\begin{tabularx}{\linewidth}[t]{@{}XR{25pt}L{24pt}R{23pt}L{16pt}@{}}
\multicolumn{5}{@{}l@{}}{\textbf{(c) HRSID}\enspace{\fontsize{7.5}{8.5}\selectfont\textcolor{gray}{mask AP, 71 methods}}} \\
\toprule
{\fontsize{7.5}{8.5}\selectfont\textcolor{gray}{\textit{Share of the training labels}}} & 100\% & & 1.2\% & \\
\midrule
\classic{Mask R-CNN}\tcite{he2017maskrcnn}\classic{\yr{2017}} & \classic{54.0} & & & \\
Insight Any Instance\tcite{li2025insight} & 59.3 & & & \\
CATNet, ms\tcite{liu2025catnet} & 59.6 & & & \\
ARANet\tcite{chang2026aranet} & 60.6 & & & \\
LFG-Net\tcite{wei2022lfgnet}\yr{2022} & \underline{63.9} & & & \\
GCIE-Net\tcite{luo2026gcienet}\yr{2026} & \underline{63.9} & & & \\
\midrule
Starter & -- &  & 36.8 &  \\
\rowcolor{eoblue!7}\method{} (GPT-6-Astra) & \textbf{68.7} & \gain{4.8} & 50.2 &  \\
\rowcolor{eoblue!7}\method{} (GPT-6.1-Sol) & \textbf{67.8} & \gain{3.9} & 50.2 &  \\
\rowcolor{eoblue!7}\method{} (Kimi-K3.1-Pro) & 59.5 &  & 45.9 &  \\
\bottomrule
\end{tabularx}
\end{minipage}\hfill
\begin{minipage}[t]{0.485\textwidth}
\begin{tabularx}{\linewidth}[t]{@{}XR{25pt}L{24pt}R{23pt}L{16pt}@{}}
\multicolumn{5}{@{}l@{}}{\textbf{(d) PASTIS}\enspace{\fontsize{7.5}{8.5}\selectfont\textcolor{gray}{mIoU, 189 methods}}} \\
\toprule
{\fontsize{7.5}{8.5}\selectfont\textcolor{gray}{\textit{Share of the training labels}}} & 100\% & & 1.4\% & \\
\midrule
\classic{U-TAE}\tcite{garnot2021pastis}\classic{\yr{2021}} & \classic{62.4} & & & \\
OlmoEarth-B\tcite{herzog2025olmoearth} & 64.3 & & & \\
TSViT\tcite{tarasiou2023tsvit} & 65.1 & & & \\
OlmoEarth-L\tcite{herzog2025olmoearth} & 66.3 & & & \\
TEA\tcite{tea2026} & 66.8 & & & \\
Exch.\,+\,M2F\tcite{cai2023exchanger}\yr{2023} & \underline{67.1} & & & \\
\midrule
Starter & 53.0 &  & 14.0 &  \\
\rowcolor{eoblue!7}\method{} (GPT-6-Astra) & \textbf{69.7} & \gain{2.6} & 50.1 &  \\
\rowcolor{eoblue!7}\method{} (GPT-6.1-Sol) & \textbf{68.3} & \gain{1.2} & 49.7 &  \\
\rowcolor{eoblue!7}\method{} (Kimi-K3.1-Pro) & 63.1 &  & 32.2 &  \\
\bottomrule
\end{tabularx}
\end{minipage}

\vspace{3pt}

\begin{minipage}[t]{0.485\textwidth}
\begin{tabularx}{\linewidth}[t]{@{}XR{25pt}L{24pt}R{23pt}L{16pt}@{}}
\multicolumn{5}{@{}l@{}}{\textbf{(e) Open-Canopy}\enspace{\fontsize{7.5}{8.5}\selectfont\textcolor{gray}{MAE (m)\,$\downarrow$, 56 methods}}} \\
\toprule
{\fontsize{7.5}{8.5}\selectfont\textcolor{gray}{\textit{Share of the training labels}}} & 100\% & & 1.5\% & \\
\midrule
\classic{UNet}\tcite{ronneberger2015unet}\classic{\yr{2015}} & \classic{2.67} & & & \\
HViT\tcite{dosovitskiy2021vit} & 2.65 & & & \\
PCPVT\tcite{chu2021twins} & 2.57 & & & \\
Swin, Satlas\tcite{bastani2023satlas} & 2.56 & & & \\
Swin\tcite{liu2021swin} & 2.54 & & & \\
PVTv2\tcite{wang2022pvtv2}\yr{2024} & \underline{2.52} & & & \\
\midrule
Starter & 4.03 &  & -- &  \\
\rowcolor{eoblue!7}\method{} (GPT-6-Astra) & \textbf{2.44} & \drop{0.08} & 2.72 &  \\
\rowcolor{eoblue!7}\method{} (GPT-6.1-Sol) & 2.70 &  & 2.73 &  \\
\rowcolor{eoblue!7}\method{} (Kimi-K3.1-Pro) & 2.72 &  & 3.04 &  \\
\bottomrule
\end{tabularx}
\end{minipage}\hfill
\begin{minipage}[t]{0.485\textwidth}
\begin{tabularx}{\linewidth}[t]{@{}XR{25pt}L{24pt}R{23pt}L{16pt}@{}}
\multicolumn{5}{@{}l@{}}{\textbf{(f) WHU-Stereo}\enspace{\fontsize{7.5}{8.5}\selectfont\textcolor{gray}{D1 (\%)\,$\downarrow$, 61 methods}}} \\
\toprule
{\fontsize{7.5}{8.5}\selectfont\textcolor{gray}{\textit{Share of the training labels}}} & 100\% & &  & \\
\midrule
\classic{PSMNet}\tcite{chang2018psmnet}\classic{\yr{2018}} & \classic{24.81} & & & \\
HMSM-Net\tcite{he2022hmsmnet} & 12.94 & & & \\
MaskCRNet\tcite{maskcrnet2024} & 12.87 & & & \\
IGEV++\tcite{xu2025igevpp} & 12.50 & & & \\
UGC-Net\tcite{ugcnet2025} & 12.19 & & & \\
Meta-MRGE\tcite{metamrge}\yr{2026} & \underline{10.00} & & & \\
\midrule
Starter & 25.61 &  &  &  \\
\rowcolor{eoblue!7}\method{} (GPT-6-Astra) & \textbf{8.95} & \drop{1.05} &  &  \\
\rowcolor{eoblue!7}\method{} (GPT-6.1-Sol) & \textbf{8.88} & \drop{1.12} &  &  \\
\rowcolor{eoblue!7}\method{} (Kimi-K3.1-Pro) & 17.31 &  &  &  \\
\bottomrule
\end{tabularx}
\end{minipage}

\vspace{3pt}

\begin{minipage}[t]{0.485\textwidth}
\begin{tabularx}{\linewidth}[t]{@{}XR{25pt}L{24pt}R{23pt}L{16pt}@{}}
\multicolumn{5}{@{}l@{}}{\textbf{(g) SEVIR}\enspace{\fontsize{7.5}{8.5}\selectfont\textcolor{gray}{CSI-M, 210 methods}}} \\
\toprule
{\fontsize{7.5}{8.5}\selectfont\textcolor{gray}{\textit{Share of the training labels}}} & 100\% & &  & \\
\midrule
\classic{ConvLSTM}\tcite{shi2015convlstm}\classic{\yr{2015}} & \classic{41.85} & & & \\
PredRNN-V2\tcite{wang2024predbench} & 43.47 & & & \\
USTEP\tcite{ustep2025} & 43.66 & & & \\
CasCast\tcite{gong2024cascast} & 44.01 & & & \\
Earthformer\tcite{gao2022earthformer} & 44.19 & & & \\
SimCast\tcite{simcast2025}\yr{2025} & \underline{45.21} & & & \\
\midrule
Starter & 38.52 &  &  &  \\
\rowcolor{eoblue!7}\method{} (GPT-6-Astra) & \textbf{47.61} & \gain{2.40} &  &  \\
\rowcolor{eoblue!7}\method{} (GPT-6.1-Sol) & \textbf{47.85} & \gain{2.64} &  &  \\
\rowcolor{eoblue!7}\method{} (Kimi-K3.1-Pro) & \textbf{45.52} & \gain{0.31} &  &  \\
\bottomrule
\end{tabularx}
\end{minipage}\hfill
\begin{minipage}[t]{0.485\textwidth}
\begin{tabularx}{\linewidth}[t]{@{}XR{25pt}L{24pt}R{23pt}L{16pt}@{}}
\multicolumn{5}{@{}l@{}}{\textbf{(h) LRS-GRO}\enspace{\fontsize{7.5}{8.5}\selectfont\textcolor{gray}{accuracy, 24 methods}}} \\
\toprule
{\fontsize{7.5}{8.5}\selectfont\textcolor{gray}{\textit{Share of the training labels}}} & 100\% & & 4.3\% & \\
\midrule
\classic{Qwen3-VL-8B}\tcite{bai2025qwen3vl}\classic{\yr{2025}} & \classic{56.9} & & & \\
AttWarp\tcite{dalal2026attwarp} & 58.2 & & & \\
ViCrop\tcite{zhang2025vicrop} & 58.8 & & & \\
WeaveEarth\tcite{ma2026weaveearth} & 58.9 & & & \\
GeoLens\tcite{wang2026geolens} & 60.7 & & & \\
GazeEarth\tcite{zhang2026gazeearth}\yr{2026} & \underline{61.0} & & & \\
\midrule
Starter & 58.1 &  & 58.1 &  \\
\rowcolor{eoblue!7}\method{} (GPT-6-Astra) & \textbf{65.7} & \gain{4.7} & \textbf{65.8} & \gain{4.8} \\
\rowcolor{eoblue!7}\method{} (GPT-6.1-Sol) & \textbf{64.8} & \gain{3.8} & \textbf{63.0} & \gain{2.0} \\
\rowcolor{eoblue!7}\method{} (Kimi-K3.1-Pro) & \textbf{63.0} & \gain{2.0} & 61.0 &  \\
\bottomrule
\end{tabularx}
\end{minipage}

\vspace{3pt}

\begin{minipage}[t]{0.485\textwidth}
\begin{tabularx}{\linewidth}[t]{@{}XR{25pt}L{24pt}R{23pt}L{16pt}@{}}
\multicolumn{5}{@{}l@{}}{\textbf{(i) LaSeRS}\enspace{\fontsize{7.5}{8.5}\selectfont\textcolor{gray}{gIoU, 30 methods}}} \\
\toprule
{\fontsize{7.5}{8.5}\selectfont\textcolor{gray}{\textit{Share of the training labels}}} & 100\% & &  & \\
\midrule
\classic{LISA-7B}\tcite{lai2024lisa}\classic{\yr{2023}} & \classic{26.6} & & & \\
GLaMM-7B\tcite{rasheed2024glamm} & 43.6 & & & \\
GeoPixel-8B\tcite{shabbir2025geopixel} & 50.4 & & & \\
SegEarth-R2\tcite{segearthr2} & 57.2 & & & \\
SELF1E-8B\tcite{self1e2026} & 68.7 & & & \\
FIRM-4B\tcite{firm2026}\yr{2026} & \underline{70.5} & & & \\
\midrule
Starter & 56.0 &  &  &  \\
\rowcolor{eoblue!7}\method{} (GPT-6-Astra) & \textbf{78.7} & \gain{8.2} &  &  \\
\rowcolor{eoblue!7}\method{} (GPT-6.1-Sol) & \textbf{71.7} & \gain{1.2} &  &  \\
\rowcolor{eoblue!7}\method{} (Kimi-K3.1-Pro) & \textbf{71.6} & \gain{1.1} &  &  \\
\bottomrule
\end{tabularx}
\end{minipage}\hfill
\begin{minipage}[t]{0.485\textwidth}
\begin{tabularx}{\linewidth}[t]{@{}XR{25pt}L{24pt}R{23pt}L{16pt}@{}}
\multicolumn{5}{@{}l@{}}{\textbf{(j) CityNav}\enspace{\fontsize{7.5}{8.5}\selectfont\textcolor{gray}{SR (\%), 48 methods}}} \\
\toprule
{\fontsize{7.5}{8.5}\selectfont\textcolor{gray}{\textit{Share of the training labels}}} & 100\% & &  & \\
\midrule
\classic{MGP}\tcite{lee2025citynav}\classic{\yr{2024}} & \classic{6.38} & & & \\
HTNav\tcite{htnav2026} & 22.23 & & & \\
GeoNav\tcite{geonav2025} & 25.90 & & & \\
CLOSER-VLN\tcite{closervln2026} & 32.01 & & & \\
ViSA\tcite{visa2026} & 36.11 & & & \\
GeoReasoner\tcite{georeasoner2026}\yr{2026} & \underline{37.77} & & & \\
\midrule
Starter & 22.54 &  &  &  \\
\rowcolor{eoblue!7}\method{} (GPT-6-Astra) & \textbf{62.38} & \gain{24.61} &  &  \\
\rowcolor{eoblue!7}\method{} (GPT-6.1-Sol) & \textbf{62.91} & \gain{25.14} &  &  \\
\rowcolor{eoblue!7}\method{} (Kimi-K3.1-Pro) & \textbf{50.52} & \gain{12.75} &  &  \\
\bottomrule
\end{tabularx}
\end{minipage}
\end{table}

\paragraph{The Median Lead Is Half of All Published Progress.} With full labels, the median lead of the agent-built models over the previous state of the art is about half of all the progress that published methods have made over the classic baseline of each task (\cref{tab:main}). The published results set a high bar. More than 1,000 methods in total report results on these tasks, and the best of them include a multi-model ensemble and fine-tuned EO foundation models. Each agent-built model comes from a single 12-hour run and improves on the starter of the harness by up to 40.4 points. The lead exceeds all of the published progress on LRS-GRO and amounts to more than two thirds of it on CityNav, SEVIR and BRIGHT. On CityNav, the agent's policy closes about half of the gap between the previous best and the 87.9\% success rate of human evaluation on the same split \cite{lee2025citynav}.

\paragraph{The Lead Holds Where the Harness Offers Least.} On LRS-GRO, the harness forbids any training, yet the pipeline the agent designs around the frozen Qwen3-VL-8B surpasses methods trained for large-image question answering as well as GazeEarth \cite{zhang2026gazeearth}, the previous best, which is training-free on the same model. On WHU-Stereo and SEVIR, the harness holds no pretrained stereo or weather-radar model, and the agent's models still surpass methods designed for each task.

\paragraph{A Single GPU and Another Developer's Agent Suffice.} On a single H100 GPU, with a quarter of the GPU hours of GPT-6-Astra, GPT-6.1-Sol surpasses the state of the art on 8 of the 10 tasks, and Kimi-K3.1-Pro, from another developer, on 6. On nine of the 10 tasks, at least two of the three agents surpass it, and on the tenth, Open-Canopy, two further runs of GPT-6-Astra do so as well, with 2.41 and 2.43\,m. A run on a full-label task costs 40 to 191 US dollars of model usage for GPT-6-Astra, 7 to 27 for GPT-6.1-Sol and 12 to 39 for Kimi-K3.1-Pro, and the 48 runs of \cref{tab:main} take up to 1,152 GPU hours (\cref{app:budget}).

\subsection{A Handful of Labels Suffices}
\label{sec:labels}

\begin{figure}[!t]
\centering
\includegraphics[width=\linewidth]{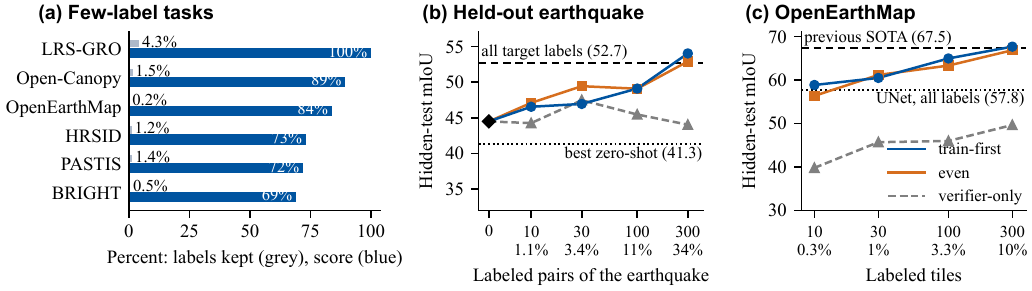}
\caption{A handful of labels suffices. (a)~Share of the training labels kept by each few-label task (grey) and of the full-label score reached by GPT-6-Astra (blue), with the inverse ratio for the MAE of Open-Canopy. (b)~Hidden-test mIoU on the earthquake held out from BRIGHT and (c)~on OpenEarthMap for three splits of the labels between training and the verifier, with GPT-6.1-Sol on a single GPU. Percentages below the budgets give their share of all labels.}
\label{fig:labels}
\end{figure}

\paragraph{Few Training Labels Recover Most of the State of the Art.} With 0.2 to 4.3\% of the training labels, the agent-built models of the six few-label tasks retain 69 to 100\% of the score of their full-label counterparts (\cref{fig:labels}a) and reach 72 to 108\% of the state-of-the-art score obtained with all labels (\cref{tab:main}). On LRS-GRO, the training-free pipeline built from 114 answered questions matches the pipeline built from all 2,656 answers and surpasses the state of the art by 4.8 points. On OpenEarthMap, six labeled tiles yield 60.0 mIoU and exceed the 57.8 of the classic UNet \cite{ronneberger2015unet} trained on the full training set.\looseness=-1

\paragraph{A Verifier Needs Only a Few Labels to Select Effective Candidates.} The label-budget tasks cap all labels, so the verifier holds only a few. On OpenEarthMap, ten labeled tiles, nine for training and one for the verifier, raise the model from 19.0 mIoU without any label to 58.9, above the classic UNet trained on all labels. With 300 tiles, 10\% of the labels and 30 of them in the verifier, the model reaches 67.7, above the previous state of the art (\cref{fig:labels}c), as do two further runs with 68.1 and 68.7. The same holds on a new disaster. On the held-out earthquake, a verifier without a single target label yields 44.5 mIoU, above every published zero-shot, domain-adaptation and semi-supervised result \cite{chen2025bright}. With 300 labeled pairs, 30 of them in the verifier, the model reaches 54.0, above the 52.7 of ChangeMamba trained on all target labels (\cref{fig:labels}b). The split of the labels matters as well. With 300 labels, keeping nine tenths for training yields the strongest model on both tasks, 67.7 and 54.0 mIoU against 49.7 and 44.0 for the verifier-only split (\cref{tab:grid}). The train-first rule thus holds on both tasks: the labels beyond the few that a verifier needs serve best in training, where they let the agent propose better candidates.

\begin{figure}[!t]
\centering
\includegraphics[width=\linewidth]{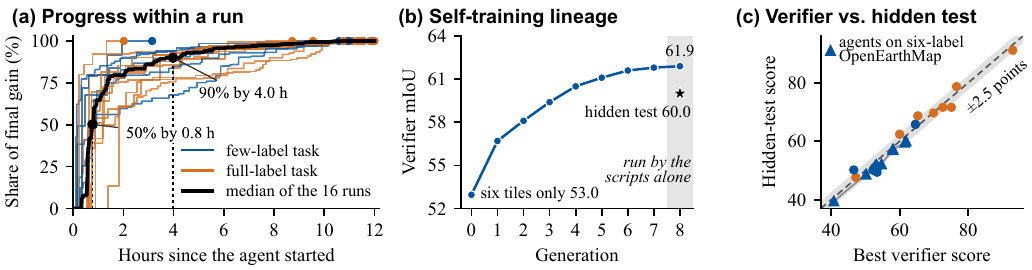}
\caption{The verifier keeps the candidates that improve the model. (a)~Share of the final verifier gain reached over time in the 16 runs of GPT-6-Astra. (b)~Verifier mIoU of the best pipeline along the self-training lineage of six-label OpenEarthMap, where generation~0 is the best model trained on the six tiles alone. (c)~Hidden-test score against best verifier score for the runs of (a) scored in points, with WHU-Stereo as $100-\mathrm{D1}$, and for the nine agents of \cref{tab:agents} (triangles), where the grey band marks a difference of at most 2.5 points. Colors in (c) follow (a).}
\label{fig:runs}
\end{figure}

\subsection{The Model Improves with the Agent}
\label{sec:agents}

\begin{table}[!t]
\begin{minipage}[t]{0.40\linewidth}
\centering
\caption{The model improves with the agent. Nine agents build models for six-label OpenEarthMap with the same data, pretrained models, verifier and 12-hour budget (\cref{app:agentsetup}). The verifier column gives the best verifier score of each run.}
\label{tab:generations}
\label{tab:agents}
\footnotesize
\setlength{\tabcolsep}{4.5pt}
\begin{tabular}{@{}llrr@{}}
\toprule
Family & Agent & Verifier & Test mIoU \\
\midrule
\multicolumn{2}{@{}l}{Starter pipeline} & -- & 34.0 \\
\midrule
\multirow{4}{*}{GPT} & GPT-5.5 & 52.9 & 51.7\gain{17.7} \\
& GPT-5.6-Sol & 54.7 & 52.2\gain{18.2} \\
& GPT-6.1-Sol & 61.5 & 59.6\gain{25.6} \\
& GPT-6-Astra & 61.9 & \textbf{60.0}\gain{26.0} \\
\midrule
\multirow{3}{*}{Kimi} & Kimi-K2.6 & 50.2 & 48.7\gain{14.7} \\
& Kimi-K3 & 57.9 & 57.0\gain{23.0} \\
& Kimi-K3.1-Pro & 58.1 & 57.2\gain{23.2} \\
\midrule
\multirow{2}{*}{Qwen} & Qwen3.8-27B & 40.8 & 39.5\gain{5.5} \\
& Qwen3.8-Max & 53.3 & 51.9\gain{17.9} \\
\bottomrule
\end{tabular}
\end{minipage}\hfill
\begin{minipage}[t]{0.57\linewidth}
\centering
\caption{The agent designs every layer of the system around the task. Examples from the runs of GPT-6-Astra.}
\label{tab:system}
\footnotesize
\setlength{\tabcolsep}{4pt}
\begin{tabularx}{\linewidth}{@{}l>{\raggedright\arraybackslash}X@{}}
\toprule
Layer & Examples \\
\midrule
Reading the data & Open-Canopy: finds that the image reader treats the near-infrared band as an alpha channel. Trains without rotations, since shadows carry height \\
Task context & HRSID: validates locally on whole held-out scenes. BRIGHT: on whole held-out events \\
Encoder & DINOv2 on six-label OpenEarthMap, DINOv3 on few-label BRIGHT, none on SEVIR and few-label PASTIS \\
Structure and metric & PASTIS: date-aware temporal attention. SEVIR: predicts threshold exceedance for CSI. HRSID: detection, then mask refinement in each box \\
Unlabeled data, ensembles & Few labels: self-training on the unlabeled pool. Full labels: ensembles \\
\bottomrule
\end{tabularx}
\end{minipage}
\end{table}

Nine agents of three families build models for six-label OpenEarthMap with the same data, pretrained models, verifier and 12-hour budget (\cref{tab:agents}). They span three generations of the GPT family, three of the Kimi family and two sizes of Qwen3.8. From the oldest to the newest generation, the model gains 7.9 mIoU with the GPT family and 8.5 with the Kimi family, Qwen3.8-Max builds a model 12.4 mIoU stronger than Qwen3.8-27B, and the flagship GPT-6-Astra builds the strongest of all. Across families, the model also improves with the general capability of the agent. Over the nine agents, its hidden-test score has a Spearman correlation of 0.87 with the Epoch Capabilities Index of the agent \cite{epoch2026eci} and of 0.90 with the rank of the agent on the LMArena WebDev leaderboard \cite{lmarena2026webdev}. The agents draw on nearly the same methods, and the stronger ones keep revising the direction of their models for longer (\cref{app:agents}).

\section{Why Agent-Built Models Surpass the State of the Art}
\label{sec:analysis}

The trajectories record every command, message and verifier query of a run. They show why agent-built models surpass the state of the art: the agent designs the whole system around each task (\cref{sec:sources}), and the verifier keeps the candidates that improve the model (\cref{sec:loop}).

\subsection{The Agent Designs the Whole System around the Task}
\label{sec:sources}

\begin{table}[!t]
\centering
\caption{Agent-built models lead the same-model baseline on all eight tasks. Each baseline is a published method that builds on a pretrained model of the harness. Scores of GPT-6-Astra follow \cref{tab:main}, with the lead in blue. $^\ast$Official code, trained on the task (\cref{app:recipes}).}
\label{tab:recipe}
\footnotesize
\setlength{\tabcolsep}{7pt}
\newcommand{\mt}[1]{\kern3pt{\fontsize{7}{8}\selectfont\textcolor{gray}{#1}}}
\begin{tabular}{@{}lllr>{\columncolor{eoblue!7}[7pt][0pt]}r@{}}
\toprule
 & \multicolumn{3}{c}{Same-model baseline} & \\
\cmidrule(lr){2-4}
Task & Method & Built on & Score & \method \\
\midrule
OpenEarthMap\mt{mIoU} & UniMatch V2$^\ast$\tcite{yang2025unimatchv2} & DINOv3-SAT-L & 69.6 & \textbf{71.6}\gain{2.0} \\
BRIGHT\mt{mIoU} & ChangeMamba\tcite{chen2024changemamba} & VMamba & 67.63 & \textbf{71.56}\gain{3.93} \\
HRSID\mt{mask AP} & Mask R-CNN$^\ast$\tcite{he2017maskrcnn} & ResNet-50 & 53.3 & \textbf{68.7}\gain{15.4} \\
PASTIS\mt{mIoU} & UPerNet$^\ast$\tcite{marsocci2026pangaea} & ResNet-50 & 37.1 & \textbf{69.7}\gain{32.6} \\
Open-Canopy\mt{MAE (m)$\downarrow$} & DPT$^\ast$\tcite{simeoni2025dinov3} & DINOv3-SAT-L & 2.89 & \textbf{2.44}\drop{0.45} \\
WHU-Stereo\mt{D1 (\%)$\downarrow$} & DEFOM-Stereo$^\ast$\tcite{jiang2025defom} & DINOv2-L & 20.90 & \textbf{8.95}\drop{11.95} \\
LRS-GRO\mt{accuracy} & GazeEarth\tcite{zhang2026gazeearth} & Qwen3-VL-8B & 61.0 & \textbf{65.7}\gain{4.7} \\
LaSeRS\mt{gIoU} & FIRM-4B\tcite{firm2026} & Qwen3-VL-4B & 70.5 & \textbf{78.7}\gain{8.2} \\
\bottomrule
\end{tabular}
\end{table}

\paragraph{The Agent Makes Opposite Choices on Different Tasks.} The agent's design reaches from how the data are read to how the predictions are combined (\cref{tab:system}), and each choice follows what the agent measures on the task. It keeps a pretrained encoder on OpenEarthMap, Open-Canopy and BRIGHT but trains temporal models from scratch on SEVIR and few-label PASTIS. On the latter, frozen DINOv3 features reach 27.2 mIoU on the verifier and the temporal models 52.2. Where labels are few, the agent self-trains on the unlabeled pool (\cref{sec:loop}).

\paragraph{The Gain Comes from the Agent's System, Not Only from the Pretrained Models.} To separate what the pretrained models of the harness contribute from what the agent adds, we compare each agent-built model with a same-model baseline, a published method that builds on a pretrained model of the harness. The agent-built model leads it on all eight tasks (\cref{tab:recipe}), by 2.0 to 32.6 points and by 0.45\,m of MAE on Open-Canopy (\cref{app:recipes,app:gains}). On OpenEarthMap, we also give the baseline the resources of the agent. Within the same 12 hours and GPUs, two runs train, select and ensemble five and seven variants of the UniMatch~V2 recipe. This fixed recipe reaches 69.8 and 70.1 mIoU, against 71.6 for the agent-built model.

\subsection{The Verifier Keeps the Candidates That Improve the Model}
\label{sec:loop}

\paragraph{Changes Follow Observations, and Most Candidates Are Discarded.} On full-label HRSID, the agent traces misplaced masks in a montage of its outputs to a half-pixel offset in its own mask refinement. Most candidates fail the agent's local checks or the verifier. On CityNav, a candidate ranker and 36,000 self-collected flights stay out of the final method (\cref{app:steps}). The candidates that the verifier keeps raise its score step by step (\cref{fig:runs}a).

\paragraph{Recursion Lifts the Six-Label Model by 9 mIoU, and the Agent Automates It.} On six-label OpenEarthMap, the agent runs eight generations of self-training (\cref{fig:teaser}b, \cref{fig:runs}b). The verifier score rises with every generation, from the 53.0 mIoU of the best model trained on the six tiles alone to 61.9. The agent then writes scripts that select the next teacher from the verifier log and start a new generation, and the scripts alone run the last one. The recursion thus extends from the model to the procedure that improves it. GPT-6-Astra runs such a loop on all five few-label tasks that allow training, with four to 13 rounds, and what passes from one generation to the next differs with the task (\cref{app:steps}).

\paragraph{Removing the Verifier's Scores or the Recursion Weakens the Model.} Nine runs of GPT-6.1-Sol on six-label OpenEarthMap, three per setting, keep both components or remove one and keep everything else. Without the scores, the verifier still runs each submitted pipeline but returns no score. Without the recursion, no network may be trained from the predictions or weights of a model trained in the run, which rules out pseudo-labels, distillation and earlier checkpoints. Over the three runs, the model reaches 59.0\,$\pm$\,0.9 mIoU on the hidden test with both components, 37.4\,$\pm$\,1.3 without the scores and 53.7\,$\pm$\,0.8 without the recursion. With the same encoder, labeled tiles and unlabeled pool, the published semi-supervised method UniMatch~V2 \cite{yang2025unimatchv2} reaches 52.4.

\paragraph{The Models Selected by the Verifier Do Not Overfit It.} Across the nine agents of \cref{tab:agents}, the verifier ranks the models exactly as the hidden test does (\cref{fig:runs}c). The hidden-test score lies within 2.5 points of the verifier score in every run on OpenEarthMap and LRS-GRO, within 0.4 points on SEVIR and full-label PASTIS and within 0.07\,m of the verifier's MAE on Open-Canopy. On HRSID, LRS-GRO and LaSeRS, it is even higher (\cref{tab:verifier}).

\section{Conclusion}
\label{sec:conclusion}

\method casts any verifiable EO task as a harness and a verifier, and these two components are all that a general-purpose coding agent needs to build the model. Within 12 hours, agent-built models surpass the state of the art on all 10 tasks of five types, and an agent on a single GPU does so on eight of them. On these tasks, the strongest model is no longer one built by hand. A handful of labels suffices, also on a new disaster, and the train-first rule gives a few of them to the verifier and the rest to training. The agent achieves these gains by designing the whole system around each task, and the verifier keeps the candidates that improve the model. Stronger agents build stronger models. A task's harness and verifier stay fixed while agents improve, so every advance in agents turns into better EO models without new human engineering. A new disaster, satellite or region then receives a model within hours, and any agency that can define a task can obtain a state-of-the-art model. We see this as the paradigm in which EO models will be built, with human effort invested in harnesses and verifiers.

\bibliographystyle{unsrtnat}
{\small
\bibliography{ref}
}

\clearpage
\appendix
\crefalias{section}{appendix}

\section*{Appendix}

This appendix details the tasks, the label budgets, the agents and the published results, lists the detailed results behind \cref{fig:labels}, and adds detail on the same-model baselines, where the gains come from, how the agents use the verifier, which steps they take, what separates the nine agents on one task and how the agents spend their budget.

\section{Tasks}
\label{app:tasks}

\paragraph{Task Descriptions.} The 10 tasks of \cref{tab:benchmarks} differ in sensor, output and source of ground truth. OpenEarthMap \cite{xia2023openearthmap} maps eight land-cover classes from sub-meter RGB imagery, BRIGHT \cite{chen2025bright} grades building damage from pre-event optical and post-event SAR pairs, HRSID \cite{wei2020hrsid} segments ship instances in SAR images, and PASTIS \cite{garnot2021pastis} maps 18 crop types from Sentinel-2 time series. Open-Canopy \cite{fogel2025opencanopy} regresses canopy height from 1.5\,m SPOT~6--7 imagery. WHU-Stereo \cite{li2023whustereo} asks for disparity maps from Gaofen-7 stereo pairs, with airborne LiDAR as ground truth. SEVIR \cite{veillette2020sevir} asks for the next hour of weather-radar echoes from the previous 65 minutes, scored against what the radar later observed. LRS-GRO \cite{liu2026zoomearth} asks questions about images of up to 27,328 pixels on a side, which the agent answers with a frozen Qwen3-VL-8B without training any parameter. LaSeRS \cite{segearthr2} asks for an ordered list of masks for a free-form query, and CityNav \cite{lee2025citynav} asks for a policy that flies a drone to a described target over city-scale aerial imagery, which the verifier runs in closed loop. OpenEarthMap, BRIGHT, HRSID, PASTIS and LaSeRS have 8, 4, 1, 19 and 122 classes, where BRIGHT and PASTIS count the background as a class. LRS-GRO forbids any training, so its labels serve only as in-context examples and for local validation.

\paragraph{Splits.} The hidden test set is the split on which published methods report their results, the official validation split of OpenEarthMap and the official test split of the other nine tasks. The verifier holds the official validation split on BRIGHT, PASTIS, Open-Canopy, WHU-Stereo and CityNav, and PASTIS uses the official fold~I. On OpenEarthMap, it holds 300 tiles from the training split, and the 1,500 official test images join the unlabeled pool. HRSID, LRS-GRO and LaSeRS have no official validation split. On HRSID, the verifier holds 20 complete scenes taken from the official training set. On LRS-GRO, it holds a quarter of the training images, and on LaSeRS, 1,600 queries split from the training set by scene. On SEVIR, the verifier holds the events from January to May 2019 and the test set those from June to November 2019. On CityNav, both the verifier and the test set fly over city blocks absent from training.

\paragraph{Few-Label Tasks.} The few-label tasks keep 6 tiles of OpenEarthMap with one per continent, 14 pairs of BRIGHT with one per event, 36 images of HRSID from 36 scenes, 20 series of PASTIS with five per tile, 1,000 tiles of Open-Canopy spread over France and 114 answers of LRS-GRO with six per question type. Together, the labels of each task cover every class, and the remaining training imagery joins the unlabeled pool.

\paragraph{Harness Contents.} The harness gives the agent what an EO researcher would start from. It contains the labeled training data, an unlabeled pool where the task provides one, the strongest general-purpose pretrained models that suit the task, the Python deep-learning and geospatial stack together with QGIS, SAGA and GDAL, and a deliberately plain starter pipeline. On the four image-interpretation tasks, Open-Canopy and WHU-Stereo, the pretrained models are encoders without decoders or task heads (DINOv3 ViT-L/16 pretrained on satellite imagery, DINOv2 ViT-L/14, ResNet-18 to ResNet-101 and EfficientNet-B0 \cite{simeoni2025dinov3,oquab2024dinov2,he2016resnet,tan2019efficientnet}) together with the official source code of the DINO models. The harness also adds the standard tools of each field, such as detection libraries for HRSID and the Swin and VMamba backbones for BRIGHT. The stereo task provides network definitions from the official repository of the benchmark but no pretrained stereo weights, and the nowcasting task provides no pretrained model. LaSeRS and CityNav provide Qwen3-VL at 2B, 4B and 8B parameters \cite{bai2025qwen3vl}, together with SAM~2.1 \cite{ravi2025sam2} for segmentation from queries and a simulator with cached Grounded-SAM detections \cite{ren2024groundedsam} for navigation. The orthophotos, the digital surface model, the landmark annotations and the cached detections are those of the CityNav benchmark and the inputs of its baseline MGP \cite{lee2025citynav}. LRS-GRO provides only Qwen3-VL-8B.

\paragraph{Starters.} The starter is an EfficientNet-B0 encoder with an FPN decoder \cite{lin2017fpn} on the four image-interpretation tasks and Open-Canopy, the official StereoNet on WHU-Stereo and a UNet on SEVIR. On LRS-GRO, it prompts Qwen3-VL-8B directly, and on LaSeRS, it takes boxes from Qwen3-VL-8B and masks from SAM~2.1 without training. On CityNav, it is a policy without a learned model that flies to the landmark matching the description and stops at its center.

\paragraph{Metrics.} Open-Canopy reports the MAE of canopy height on vegetated pixels below 60\,m, with the original LiDAR labels of \citet{fogel2025opencanopy}. On WHU-Stereo, D1 is the share of valid pixels whose disparity is off by more than 3 pixels. SEVIR reports CSI-M, the critical success index averaged over six VIL thresholds, on the 12,159 test sequences, where a pipeline writes its 12 forecast frames in VIL levels and the grader applies the thresholds. LRS-GRO reports question-level accuracy on the 9,734 test questions under the official judge, and LaSeRS reports gIoU averaged over its nine subsets on the public test set. CityNav reports the success rate on 5,311 test episodes, where an episode counts as a success when the drone stops within 20\,m of the target, as in the protocol of \citet{lee2025citynav}.

\section{Label Budgets}
\label{app:budgets}

On OpenEarthMap and on an earthquake held out from BRIGHT, a task holds 0, 10, 30, 100 or 300 labeled samples, and larger budgets contain the smaller ones. Three splits divide them between the harness and the verifier. The train-first split gives nine tenths to training, the even split half, and the verifier-only split all of them to the verifier. On OpenEarthMap, the labels come from the 300 verifier tiles of the full-label task, and the 2,700 training tiles join the unlabeled pool without labels. On BRIGHT, the 2023 Turkey earthquake, its largest event, is held out as in the cross-event protocol of \citet{chen2025bright}. The agent trains on the labels of the other 13 events, the labeled pairs come from the earthquake, its remaining imagery outside the test set joins the unlabeled pool, and the hidden test set holds its 224 test pairs. Without target labels, the verifier holds pairs of the source events.

\paragraph{A Verifier Needs Only a Few Labels.} The few-label tasks of \cref{app:tasks} limit the training labels, and their verifiers keep the validation split of \cref{tab:benchmarks}. The label-budget tasks cap the labels of training and of the verifier together. On OpenEarthMap, GPT-6.1-Sol reaches 58.9 mIoU with nine training tiles and a single verifier tile (\cref{tab:grid}), close to the 59.0\,$\pm$\,0.9 it reaches with six training tiles and the 300-tile verifier (\cref{sec:loop}). With 270 training tiles and 30 verifier tiles, about as many labels in total as the six training tiles and the 300-tile verifier of the six-label task, it reaches 67.7, 8.7 mIoU higher.

\paragraph{Detailed Results.} \Cref{tab:grid} lists the scores behind \cref{fig:labels}b and c. On the held-out earthquake, \cref{tab:disaster} lists the published results that \cref{sec:labels} compares with, and the model of every budget and split surpasses the best published zero-shot, domain-adaptation, semi-supervised and one-shot results. The verifier-only split trails the better of the other two at every budget from 10 to 300 labels, by 15 to 19 mIoU on OpenEarthMap and by 2.0 to 10.0 on the held-out earthquake.

\begin{table}[!htb]
\centering
\caption{Hidden-test mIoU of the models built by GPT-6.1-Sol on a single H100 GPU within 12 hours as a task holds more labels, split between training and the verifier. The train-first split gives nine tenths of the labels to training, the even split half, and the verifier-only split all of them to the verifier. Without target labels, the BRIGHT verifier holds pairs of the source events, and the OpenEarthMap task holds no label at all.}
\label{tab:grid}
\footnotesize
\setlength{\tabcolsep}{6pt}
\begin{tabular}{@{}rrrrrrr@{}}
\toprule
 & \multicolumn{3}{c}{BRIGHT, held-out earthquake} & \multicolumn{3}{c}{OpenEarthMap} \\
\cmidrule(lr){2-4}\cmidrule(l){5-7}
Labels & train-first & even & verifier-only & train-first & even & verifier-only \\
\midrule
0 & \multicolumn{3}{c}{44.48} & \multicolumn{3}{c}{19.00} \\
10 & 46.54 & 47.09 & 44.25 & 58.86 & 56.36 & 39.79 \\
30 & 46.94 & 49.43 & 47.46 & 60.49 & 61.17 & 45.69 \\
100 & 49.09 & 49.08 & 45.46 & 64.97 & 63.33 & 45.97 \\
300 & 54.03 & 52.90 & 44.03 & 67.69 & 66.86 & 49.65 \\
\bottomrule
\end{tabular}
\end{table}

\begin{table}[!htb]
\centering
\caption{Published results of \citet{chen2025bright} on the held-out 2023 Turkey earthquake of BRIGHT and the models built by GPT-6.1-Sol on a single H100 GPU within 12 hours (\cref{tab:grid}). All report 4-class mIoU on the 224 test pairs of the earthquake. The published result with all target labels comes from the standard split, whose training set holds the training pairs of all 14 events.}
\label{tab:disaster}
\footnotesize
\setlength{\tabcolsep}{6pt}
\begin{tabular}{@{}lr@{}}
\toprule
Setting & mIoU \\
\midrule
Best published zero-shot, ChangeMamba \cite{chen2024changemamba} & 41.28 \\
Best published domain adaptation, FDA \cite{yang2020fda} & 36.82 \\
Best published semi-supervised, CCT \cite{ouali2020cct} & 39.72 \\
Best published one-shot, ChangeMamba \cite{chen2024changemamba} & 42.80 \\
Published, all target labels, ChangeMamba \cite{chen2024changemamba} & 52.70 \\
\midrule
\method, no target label & 44.48 \\
\method, 300 target labels, 30 of them in the verifier & 54.03 \\
\bottomrule
\end{tabular}
\end{table}

\section{Agents and Protocol}
\label{app:agentsetup}

Three agents run every full-label task and the six few-label tasks, each in its own coding interface: GPT-6-Astra in Codex on four GB200 GPUs, and GPT-6.1-Sol in Codex and Kimi-K3.1-Pro in Kimi Code, each on a single H100 GPU. GPT-6.1-Sol in Codex also runs the label-budget tasks, each on a single H100 GPU. On six-label OpenEarthMap, GPT-5.5, GPT-5.6-Sol and GPT-6.1-Sol in Codex and Kimi-K2.6, Kimi-K3 and Kimi-K3.1-Pro in Kimi Code each run on a single H100 GPU, GPT-6-Astra runs on four GB200 GPUs as in the main results, and Claude Code runs Qwen3.8-Max and Qwen3.8-27B through the Anthropic-compatible interface of OpenRouter on four GB200 GPUs. All agents use their highest reasoning effort and work for at most 12 hours in an offline container (\cref{sec:harness}), and no single training run may exceed 4 hours. The runs took place in October and November 2026, with Codex 0.162 and Kimi Code 2.4.1.

The container of the agent holds no input or label of the verifier or of the hidden test set, and the model is fixed when it is scored. When the verifier or the grader runs a pipeline, the pipeline only infers and may not learn at test time, whether by updating on the evaluation inputs or by using one input sample to predict another. Nothing else the agent created survives, so every model, weight and script the final pipeline needs must be inside the submission.

\paragraph{Task Instruction.} Every task gives the agent an instruction of the same form, which states the task and its metric, the contents of the harness, the rules above and how to query the verifier. It gives no published score and prescribes no method, and it permits methods such as pseudo-labeling on the unlabeled pool without requiring any (\cref{sec:recursion}).

\section{Published Results}
\label{app:published}

All published results in \cref{tab:main} use the standard split and metric of their task. OpenEarthMap scores are 8-class means recomputed from the per-class IoUs, and PASTIS uses fold~I. The LRS-GRO baselines are as reported by \citet{zhang2026gazeearth}, and the LaSeRS baselines as reported by \citet{segearthr2} and \citet{firm2026}. The method counts in \cref{tab:main} cover the methods that report a result on the task itself, as of 8 November 2026. Some of them use the following settings.
\begin{compactitem}
\item UNetFormer, MetaSegNet and PyramidMamba: test-time augmentation.
\item Ensemble on BRIGHT: ensemble with test-time augmentation and building-level voting.
\item LFG-Net: ResNeXt-64$\times$4d backbone with multi-scale training.
\item OlmoEarth-B and OlmoEarth-L: fine-tuned, as reported by \citet{herzog2025olmoearth}.
\item U-TAE and Exchanger with Mask2Former: on whole $128\times128$ patches, from the released U-TAE weights and the released Exchanger test logs \cite{cai2023exchangerweights}.
\item TSViT and TEA: on $24\times24$ crops \cite{tarasiou2023tsvit,tea2026}.
\item IGEV++: retrained by \citet{metamrge}.
\item Qwen3-VL-8B, AttWarp, ViCrop, WeaveEarth and GazeEarth: training-free on Qwen3-VL-8B, as are all agent models on LRS-GRO.
\item SELF1E-8B: retrained on LaSeRS by \citet{firm2026}.
\end{compactitem}

\paragraph{The Median Lead Is 54\% of the Published Progress.} The classic baseline of a task is its standard architecture, with the score of the paper that introduced the dataset or its protocol where it reports one. The year of a classic baseline is that of the method, and the year of a previous state of the art is that in which the result first appeared. The published progress of a task (\cref{sec:results}) is the difference between its previous state of the art and its classic baseline, and the lead of the strongest agent-built model of \cref{tab:main} amounts to 54\% of it at the median of the 10 tasks, as does the lead of GPT-6-Astra alone.

\section{Same-Model Baselines}
\label{app:recipes}

A same-model baseline is a published method that builds on a pretrained model of the harness, which separates what the pretrained models contribute from what the agent adds (\cref{sec:sources}). Three same-model baselines of \cref{tab:recipe}, ChangeMamba, GazeEarth and FIRM-4B, carry their published scores (\cref{tab:main}). The final full-label pipeline of each of the three tasks includes the pretrained model that the baseline builds on: VMamba among the encoders of its ensemble on BRIGHT, the frozen Qwen3-VL-8B on LRS-GRO and Qwen3-VL-4B in two of its three LoRA-tuned localizers on LaSeRS. On OpenEarthMap, the baseline is the supervised recipe of UniMatch~V2 \cite{yang2025unimatchv2}, which fine-tunes the satellite-pretrained DINOv3 ViT-L with a DPT head \cite{ranftl2021dpt}. The agent-built model uses the same encoder (\cref{tab:gains}). For the comparison of \cref{sec:loop}, UniMatch~V2 is trained with its official semi-supervised code and the satellite-pretrained DINOv3 ViT-L on the six labeled tiles and the unlabeled pool of six-label OpenEarthMap and is graded once. The final ensembles of GPT-6.1-Sol keep this encoder as well.

The baselines on HRSID, PASTIS, Open-Canopy and WHU-Stereo and the supervised UniMatch~V2 use the official code and default hyperparameters of the published method, train on the full training set with a single seed, select the checkpoint by its verifier score and are graded once by the grader of the task, without ensembling or test-time augmentation. On HRSID, the baseline is Mask R-CNN with a ResNet-50 FPN backbone in MMDetection \cite{chen2019mmdetection}. On PASTIS, it is the ResNet-50 encoder with L-TAE temporal encoding \cite{garnot2020ltae} and an UPerNet decoder \cite{xiao2018upernet} of PANGAEA \cite{marsocci2026pangaea}, on all time steps. On Open-Canopy, it is the frozen satellite-pretrained DINOv3 ViT-L with a DPT head. On WHU-Stereo, it is DEFOM-Stereo \cite{jiang2025defom}, trained for at most 10,000 steps with a frozen DINOv2-L encoder and the official Depth Anything V2 Large weights \cite{yang2024depthanythingv2} for its depth head, which the harness does not provide to the agent.

\section{Where the Gains Come From}
\label{app:gains}

\begin{table}[!htb]
\centering
\caption{Sources of the verifier gain in 14 of the 16 runs of GPT-6-Astra. Each improvement of the best verifier score is attributed to the change the agent made at that step, from its notes and code. Shares refer to the gain from the starter to the final verifier score. On full-label BRIGHT, PASTIS and HRSID, the first submission already includes the encoder, and the shares give what the agent adds on top of it. Shares are rounded.}
\label{tab:gains}
\footnotesize
\setlength{\tabcolsep}{3.6pt}
\resizebox{\linewidth}{!}{%
\begin{tabular}{@{}llllrrr@{}}
\toprule
Task & Labels & Largest source & Second source & Ens., TTA & Self-train. & Calib. \\
\midrule
OpenEarthMap & few & encoder (DINOv2), 63\% & pseudo-labels, 8 rounds, 31\% & 2\% & 31\% & 2\% \\
OpenEarthMap & full & encoder (DINOv3), 71\% & head 9\%, recipe 9\% & 10\% & 0 & 0 \\
BRIGHT & few & encoder (DINOv3), 38\% & self-training 19\%, building-level stages 18\% & 13\% & 19\% & 3\% \\
BRIGHT & full & ensemble, 43\% & recipe 28\%, event-aware head 25\% & 43\% & 0 & 5\% \\
HRSID & few & self-training, 5 rounds, 34\% & box-cropped mask refinement, 33\% & 22\% & 34\% & 3\% \\
HRSID & full & half-pixel alignment fix, 31\% & ensemble and TTA 29\%, architecture 17\% & 29\% & 0 & 7\% \\
PASTIS & few & temporal architecture, 39\% & unlabeled data 25\%, recipe 23\% & 4\% & 25\% & 7\% \\
PASTIS & full & ensemble, 55\% & calibration 13\%, recipe 12\% & 55\% & 0 & 13\% \\
WHU-Stereo & full & learned per-pair disparity offset, 70\% & architecture 10\%, encoder 8\% & 7\% & 0 & 1\% \\
SEVIR & full & recipe, 35\% & ensemble 35\%, architecture 17\% & 35\% & 0 & 10\% \\
LRS-GRO & few & zoom-in crops at high resolution, 45\% & in-context examples, 30\% & 20\% & -- & 5\% \\
LRS-GRO & full & zoom-in crops at high resolution, 45\% & in-context examples, 35--40\% & 15\% & -- & $\le$5\% \\
LaSeRS & full & LoRA-tuned localization, 72\% & pipeline, 11\% & 9\% & -- & 1\% \\
CityNav & full & LoRA-tuned VLM localizer, 37\% & per-episode visual memory 32\%, landmark prior 23\% & 0 & 0 & 4\% \\
\bottomrule
\end{tabular}}
\end{table}

In five of the 16 runs of GPT-6-Astra, the largest source of the verifier gain is a pretrained encoder, and in the other eleven, it is the agent's own design, such as a temporal architecture, an ensemble or zoom-in crops. \Cref{tab:gains} attributes each improvement of the verifier score to the change the agent made at that step for 14 of these runs, and on Open-Canopy, the DINOv3 encoder accounts for most of the gain of the few-label run, and the encoder and longer training together account for about 85\% of the gain of the full-label run. A new encoder with its decoder accounts for 38 to 85\% of the gain on OpenEarthMap, Open-Canopy and few-label BRIGHT, self-training on the unlabeled pool for 19 to 34\% with few labels, and ensembles and test-time augmentation for 29 to 55\% on full-label BRIGHT, HRSID, PASTIS and SEVIR. On LRS-GRO, the agent locates the region a question refers to and crops it from the full-resolution image, which accounts for about 45\% of its verifier gain. On WHU-Stereo, it observes that the disparities of a pair are shifted by a nearly constant offset and trains a network on the training pairs to predict this offset from the two images of a pair, which accounts for about 70\% of its verifier gain. On few-label PASTIS, the agent tries the pretrained encoders of the harness and keeps temporal models trained from scratch instead. Frozen DINOv3 features score 27.2 mIoU on the verifier, against 52.2 for the final temporal models.

\section{How Agents Use the Verifier}
\label{app:verifieruse}

\begin{table}[!htb]
\centering
\caption{Verifier and hidden-test scores of the runs of GPT-6-Astra, grouped by whether the agent built a local validation set from its labels. The gap is the best verifier score minus the hidden-test score in points. It is computed from the unrounded scores and can differ in its last digit from the difference of the rounded entries. For Open-Canopy, the scores are the MAE in meters, 2.435 and 2.723\,m on the hidden test before rounding, and the gap is the test MAE minus the verifier MAE. On WHU-Stereo, the scores are $100-\mathrm{D1}$. On LaSeRS, the test score averages nine subsets and the verifier score all queries.}
\label{tab:verifier}
\footnotesize
\setlength{\tabcolsep}{4.5pt}
\begin{tabular}{@{}lllrrr@{}}
\toprule
Task & Agent & Local validation & Verifier & Test & Gap \\
\midrule
\multicolumn{6}{@{}l}{\textit{With a local validation set}} \\
PASTIS, full & GPT-6-Astra & 160 series, then five folds & 69.9 & 69.7 & $+$0.2 \\
OpenEarthMap, full & GPT-6-Astra & 170 tiles & 72.6 & 71.6 & $+$1.0 \\
HRSID, full & GPT-6-Astra & 301 images, by scene & 65.3 & 68.7 & $-$3.4 \\
BRIGHT, full & GPT-6-Astra & 148 pairs, by event & 75.1 & 71.6 & $+$3.5 \\
Open-Canopy, full & GPT-6-Astra & 768 tiles & 2.36 & 2.44 & $+$0.07 \\
Open-Canopy, few & GPT-6-Astra & 100 tiles, by height & 2.68 & 2.72 & $+$0.05 \\
LRS-GRO, full & GPT-6-Astra & 635 questions & 64.6 & 65.7 & $-$1.2 \\
LRS-GRO, few & GPT-6-Astra & 114 questions & 64.6 & 65.8 & $-$1.2 \\
WHU-Stereo, full & GPT-6-Astra & one city & 93.0 & 91.1 & $+$2.0 \\
SEVIR, full & GPT-6-Astra & 405 events & 47.2 & 47.6 & $-$0.4 \\
LaSeRS, full & GPT-6-Astra & 902 queries & 76.6 & 78.7 & $-$2.1 \\
CityNav, full & GPT-6-Astra & 1,500 flights, by scene & 60.0 & 62.4 & $-$2.3 \\
\midrule
\multicolumn{6}{@{}l}{\textit{Verifier as the only validation signal}} \\
PASTIS, few & GPT-6-Astra & none & 52.2 & 50.1 & $+$2.1 \\
OpenEarthMap, few & GPT-6-Astra & none & 61.9 & 60.0 & $+$2.0 \\
HRSID, few & GPT-6-Astra & none & 46.6 & 50.2 & $-$3.5 \\
BRIGHT, few & GPT-6-Astra & none & 53.4 & 49.4 & $+$3.9 \\
\bottomrule
\end{tabular}
\end{table}

\paragraph{Local Validation.} Like an EO researcher, GPT-6-Astra builds a validation signal of its own from the labels it holds in 12 of its 16 runs (\cref{tab:verifier}). It holds out whole scenes on HRSID and whole events on BRIGHT and uses five folds on PASTIS. It then screens candidates locally before it submits them to the verifier. In the four few-label runs on PASTIS, OpenEarthMap, HRSID and BRIGHT, it relies on the verifier alone. The verifier score and the hidden-test score differ in both directions, and in 7 of the 16 runs, the hidden-test score is the better of the two.

\section{Steps and Self-Training Loops}
\label{app:steps}

The trajectories show how the agent designs the system around each task: its steps fall into a few recurring kinds, new designs follow from what it observes, and its self-training loops differ with the task. \Cref{tab:system} gives examples at every layer of the system that the agent designs.

\begin{compactitem}
\item \textbf{Choosing the Encoder.} Where it keeps a pretrained encoder, the agent replaces the starter within about 20 minutes of a run with the encoder that the verifier favors and a decoder of its own. This first model raises the verifier score by 16.1 mIoU on six-label OpenEarthMap and by 7.3 mIoU on few-label BRIGHT and lowers the verifier MAE on few-label Open-Canopy from 3.90 to 3.02\,m. The choice follows the task. The verifier favors DINOv2 on six-label OpenEarthMap and the satellite-pretrained DINOv3 on few-label BRIGHT and Open-Canopy.
\item \textbf{Matching Training to the Labels.} With a handful of labels, the agent learns from the verifier that brief training generalizes better than long training, and on few-label Open-Canopy, it fine-tunes only the last 12 of the 24 encoder blocks.
\item \textbf{Debugging.} In the full-label HRSID run, the agent finds that mixed precision shifts box coordinates by several pixels, and from a montage of its outputs, it traces misplaced masks to a half-pixel offset in mask refinement (\cref{sec:loop}).
\item \textbf{Task-Specific Structure.} On BRIGHT with few labels, the agent separates the confidence that a pixel is a building from the confidence in its damage grade when it trains on pseudo-labels, which adds 2.2 points in a controlled comparison. A separate optical detector for building footprints adds a further 1.1 points. On LRS-GRO, it crops the region a question refers to at high resolution, adds answered training questions as in-context examples and votes over several answers. Besides training without rotations and flips on few-label Open-Canopy, it enlarges the input so that each ViT patch covers 12\,m instead of 24\,m.
\end{compactitem}

\paragraph{Observations and Discarded Candidates.} New designs follow from what the agent observes, and its local checks and the verifier keep those that improve the model. On CityNav, it computes locally that one of its six candidate locations lies within 20\,m of the target in 72.5\% of episodes, whereas its policy selects the right one in 60.3\%, and only then tries a ranker for the candidates. On LRS-GRO, it notices that a change lowers the accuracy on relative-position questions and reverts it. Most proposals do not survive the agent's local checks or the verifier. On CityNav, the candidate ranker, a heatmap head and 36,000 self-collected flights over the training scenes stay out of the final method, and on few-label BRIGHT, VMamba, LoRA and EMA are dropped.

\paragraph{The Loops Differ with the Task.} The few-label runs of GPT-6-Astra run eight generations of self-training on OpenEarthMap, ten on BRIGHT, five on HRSID, 13 rounds on PASTIS and four on Open-Canopy. On six-label OpenEarthMap, GPT-6.1-Sol runs such a loop as well, with an online and an offline teacher. The loops differ in what passes from one generation to the next. On BRIGHT, the teacher is a three-stage pipeline of a building detector, a damage network and a building-level classifier, and its first two generations add about 2 and 1 points. On HRSID, the teacher is a detector followed by a network that predicts each mask from a crop around the detected box. On Open-Canopy, the agent weights the pseudo-labels of 65,339 unlabeled tiles by the uncertainty of its teacher.

\section{Where Stronger Agents Gain}
\label{app:agents}

The nine runs of \cref{tab:agents} share the task, the verifier and the 12-hour budget. We label each submission of these runs by the level of the change it makes relative to the previous submission. A direction-level change replaces the encoder, switches between a frozen and a fine-tuned encoder, replaces the decoder or uses pseudo-labels for the first time. A recipe-level change starts a new round of pseudo-labels or changes thresholds, learning rates, losses or seeds, and an inference-level change alters the ensemble, test-time augmentation, calibration or class biases (\cref{tab:levels}).

\begin{table}[!htb]
\centering
\caption{Levels of change along the nine runs of \cref{tab:agents}. Direction-level changes count the submissions that change the direction of the model, the next column gives the time of the last one, and the inference share is the share of submissions in the first two hours that change only inference or the ensemble. The Qwen3.8 agents submit nothing in their first two hours.}
\label{tab:levels}
\footnotesize
\setlength{\tabcolsep}{4pt}
\begin{tabular}{@{}lrrrr@{}}
\toprule
Agent & Test mIoU & Direction & Last direction (h) & Inference share \\
\midrule
GPT-6-Astra & 60.0 & 19 & 6.8 & 22\% \\
GPT-6.1-Sol & 59.6 & 16 & 9.4 & 15\% \\
Kimi-K3.1-Pro & 57.2 & 5 & 2.8 & 67\% \\
Kimi-K3 & 57.0 & 3 & 2.7 & 33\% \\
GPT-5.6-Sol & 52.2 & 5 & 3.0 & 50\% \\
Qwen3.8-Max & 51.9 & 3 & 5.5 & -- \\
GPT-5.5 & 51.7 & 3 & 1.7 & 53\% \\
Kimi-K2.6 & 48.7 & 8 & 2.4 & 38\% \\
Qwen3.8-27B & 39.5 & 1 & 5.5 & -- \\
\bottomrule
\end{tabular}
\end{table}

\paragraph{The Agents Draw on Nearly the Same Methods.} Almost every run uses a pretrained DINO encoder, pseudo-labels on the unlabeled pool, an ensemble and test-time augmentation, and even Qwen3.8-27B, whose model is the weakest of the nine, writes a pseudo-labeling script within 1.2 hours.

\paragraph{Where the Runs Diverge.} GPT-6-Astra and GPT-6.1-Sol keep changing the direction of their models late into the run, with 19 and 16 direction-level changes, the last after 6.8 and 9.4 hours, and spend 22 and 15\% of their early submissions on inference and ensembling. The other GPT and Kimi agents make 3 to 8 direction-level changes, the last within 3.0 hours, and spend 33 to 67\% of their early submissions on inference and ensembling. Their directions often rest on a single comparison. GPT-5.6-Sol, for instance, keeps its encoder frozen for the rest of its run after fine-tuning loses 0.4 mIoU in one comparison made before any pseudo-labels exist.

\paragraph{Recursion Overturns Early Conclusions.} Each direction is chosen while the agent holds only six labels, but pseudo-labels change the training data and with it the ranking of the options. With a frozen encoder, DINOv2 leads the satellite-pretrained DINOv3 by 4 to 7 mIoU, whereas with pseudo-labels, the two tie in two runs of GPT-6.1-Sol, at 58.2 against 57.8 and 55.0 against 54.9 verifier mIoU, and fine-tuning the encoder turns from harmful to helpful. GPT-6.1-Sol keeps frozen, LoRA and fully fine-tuned routes on both encoders alive until the end of the run, and on four GPUs, GPT-6-Astra returns to the frozen encoder after 4.95 hours and adds the resulting model to its final ensemble. Agents that re-test earlier decisions under the new training data keep improving, whereas agents that treat them as settled refine a direction that is no longer the best.

\section{Budget, Stopping and Cost}
\label{app:budget}

\paragraph{A Strong Model within Hours.} Across the 16 runs of GPT-6-Astra, half of the final verifier gain, measured from the first to the best verifier score of a run, is reached after a median of 0.8 hours and 90\% after 4.0 hours (\cref{fig:runs}a). In both LRS-GRO runs, the agent declares its pipeline complete and stops after 2.0 and 3.1 hours. The 12-hour budget is therefore an upper bound, and a model close to the final one is available within the first few hours, which is the time scale that rapid response requires.

\paragraph{Cost.} The model-usage cost follows from the token counts in the trajectory logs and the list price of each model. A run of GPT-6-Astra costs about 40 US dollars on LRS-GRO and 107 to 191 US dollars on the other tasks. On the 10 full-label tasks, a run of GPT-6.1-Sol costs 7 to 27 US dollars and a run of Kimi-K3.1-Pro 12 to 39. The four GB200 GPUs of GPT-6-Astra and the 12-hour budget cap the compute of a run at 48 GPU hours, and a single H100 GPU at 12 GPU hours.

\end{document}